# Predicting the Stay Length of Patients in Hospitals using Convolutional Gated Recurrent Deep Learning Model

Mehdi Neshat[1,2], Michael Phipps[1], Chris A. Browne[3], Nicole T. Vargas[4], Seyedali Mirjalili[2]

[1] Canberra Health Services, Canberra Hospital, Canberra, ACT 2605, Australia

[2] Center for Artificial Intelligence Research and optimisation, Torrens University Australia, Brisbane, QLD 4006, Australia
neshat.mehdi@torrens.edu.au, ali.mirjalili@torrens.edu.au

[3] Deputy Vice-Chancellor (Academic) Portfolio, The Australian National University, Canberra, ACT 2605, Australia chris.browne@anu.edu.au

[4] School of Medicine and Psychology, College of Health and Medicine, Australian National University, 54 Mills Rd, Florey Building, Canberra, ACT 2601, Australia
Nicole.Vargas@anu.edu.au

## Abstract

Predicting hospital length of stay (LoS) stands as a critical factor in shaping public health strategies. This data serves as a cornerstone for governments to discern trends, patterns, and avenues for enhancing healthcare delivery. In this study, we introduce a robust hybrid deep learning model, a combination of Multi-layer Convolutional (CNNs) deep learning, Gated Recurrent Units (GRU), and Dense neural networks, that outperforms 11 conventional and state-ofthe-art Machine Learning (ML) and Deep Learning (DL) methodologies in accurately forecasting inpatient hospital stay duration.

Our investigation delves into the implementation of this hybrid model, scrutinising variables like geographic indicators tied to caregiving institutions, demographic markers encompassing patient ethnicity, race, and age, as well as medical attributes such as the CCS diagnosis code, APR DRG code, illness severity metrics, and hospital stay duration. Statistical evaluations reveal the pinnacle LoS accuracy achieved by our proposed model (CNN-GRU-DNN), which averages at 89% across a 10-fold cross-validation test, surpassing LSTM, BiLSTM, GRU, and Convolutional Neural Networks (CNNs) by 19%, 18.2%, 18.6%, and 7%, respectively.

Accurate LoS predictions not only empower hospitals to optimise resource allocation and curb expenses associated with prolonged stays but also pave the way for novel strategies in hospital stay management. This avenue holds promise for catalysing advancements in healthcare research and innovation, inspiring a new era of precision-driven healthcare practices.

## 1. Introduction

One of the significant factors in managing hospital expenses, showing the efficiency of the hospital care unit and patient flow, is the length of stay of patients (LoS) during a single visit event. Therefore, predicting the stay length of patients with high accuracy has made an interesting field of investigation in improving health services. This is mainly because indicating LoS can result in [1]: minimising hospital costs, improving patient care, and enhancing health service efficiency. Developing a robust prediction model for estimating the stay length of patients in hospitals with high generalisation capability for extending this efficient model across several hospital environments can be challenging. This is mainly because of the complex nature of treatments; for instance, the LoS for the same disease can be varied from 2 to more than 50 days. The main reasons for this significant variation may come from the characteristics of patients, the uncertainty and complexity of treatment, and also social circumstances (such as marital status, religion, healthy family relationships/availability, gender, location of patient residence, etc).

The nonlinear relationship between the dynamic environment of hospitals and patient flows makes a complex system with high uncertainty. Therefore, developing an accurate model for estimating the LoS should be able to deal with these listed challenges. Data methodologies within the Length of Stay (LoS) prediction domain are commonly classified into two main groups: classification models and regression models, as referenced in the available works of literature [2, 3]. Classification models in this context aim to categorise LoS into various groupings like short, moderate, and extended stays, aligning with the duration of patient hospitalisations in terms of days. Various methods have been used to model the complexity and uncertainty associated with this problem's nature in order to predict the LoS. Generally speaking, in AI methods, the characteristics of patients and environmental factors are used as features, and the stay length of patients in hospitals is used as the target. In order to evaluate the relationship between different administrative features like date of discharge, age, and comorbidities (having more than one illness at once) with LoS [4].

Some AI methods, such as ANNs, SVMs, and deep learning, have demonstrated a high ability to classify the LoS with considerable accuracy [5]. One recent study [6] proposed a Classifier Fusion-LoS for estimating the length of stay for infants in the Neonatal Intensive Care Unit (NICU). The study conducted two-class predictions for short- and long-term stays. Fusion-LoS is composed of an ensemble voting classifier for outperformed other classical models such as Logistic Regression (LR), Random Forest (RF), KNN, and Support Vector Classifier (SVC), as well as AdaBoost, XGBoost, and CatBoost, and achieved the highest validation accuracy of 0.96%. Another study [7] focused on classification type, employed various machine learning techniques to predict hospital procedure lengths of stay (LOS) and developed models for newborns and non-newborn groups. Results showed predictive solid performance with notable R2 scores, such as 0.82 for newborns using linear regression and 0.43 for non-newborns with Catboost regression and focusing on cardiovascular disease improved prediction to an R2 score of 0.62. In classification tasks, Multinomial Logistic Regression achieved 46.98% accuracy for non-newborns, while Random Forest Classification reached 60.08% for newborns. Roy et al. [8] compared random forests and neural networks in two scenarios, incorporating different numbers of features in order to classify consultation length in Emergency Departments (EDs), and reported an accuracy of 63.4% and 29.6%, respectively. This classifier suggested that consultation time hinges on various factors like arrival mode, symptoms, age, doctor's ID, and ED queue size and finally, Random forests outperformed neural networks in testing accuracy, offering simplicity and interpretability for deriving consultation time policies. One example of convolutional neural networks is a study that used temporal CNN models [9] with data re-balancing to predict the patient's length of stay and mortality in real-time. The proposed TCNN outperformed baselines in mortality tasks and competes well in LoS tasks. Moreover, re-balancing improved predictive power with TCNN offers efficiency and strong performance in mortality classification. The research suggested including TCNs in critical care outcome prediction systems.

While the existing literature emphasises the importance of using classification models to predict LoS, recent research has drawn attention to the significantly right-skewed patterns in LoS distributions [10]. This skewness often results in an imbalance within the dataset [11], with very few occurrences of extended LoS, which can affect the model's performance evaluation. In many cases, long LoS instances are mistakenly considered outliers [12], further complicating the analysis. As a result, many experts suggest interpreting LoS prediction as a regression task to provide more suitable and insightful results for healthcare decision-making. Three decision tree regression models, including bagging, AdaBoost, and RF, were trained by Ma et al. [12] in order to make predictions regarding the duration of stay (LoS) for hospitalised individuals. The bagging technique stood out among the three models, exhibiting the most promising outcomes during testing, showcasing a remarkably low root mean square error (RMSE) of 0.296, a high R2 value of 83.1%, and an impressive accuracy rate of 72.3%. They [12] not only trained the decision tree regression models but also delved into the intricacies of each method, shedding light on the strengths and weaknesses of bagging, AdaBoost, and RF in LoS prediction. Another group of researchers, Hasan et al. [13], delved into LoS prediction utilising the MIMIC-III dataset and tested five different ML algorithms. Among these algorithms, the extreme gradient boosting (XGBoost) regressor emerged as the top performer, boasting an RMSE of 1.2 and an R2 value of 86%. Similarly, Boff et al. [14] explored the realm of LoS forecasting by employing five diverse AI-based algorithms, namely multiple linear regression, RF, SVR, ridge regression, and partial least squares algorithms. The RF model shone brightly in their experimentation, demonstrating a notable R2 value of 65.7% and a mean absolute error (MAE) of 3.51 days. The realm of predicting hospital stays has seen a surge in interest from researchers, with various innovative approaches being explored to enhance the accuracy of such predictions.

Recent studies have increasingly focused on improving Length of Stay (LoS) predictions by utilising a range of medical data types (heterogeneous) [15], such as tabular clinical notations, laboratory outcomes, bio-signals, medical images and demographic details. For example, Chen et al. [16] proposed employing multi-modal ML algorithms to provide accurate LoS predictions based on demographic data and medical chest X-ray images. Their approach achieved R2 and EVAR values of 60.4% and 0.6, surpassing competitors in LoS prediction on the Medical Information Mart for Intensive Care (MIMIC)-IV test dataset. In another similar work in terms of multi-modal datasets, Zhang et al. [17] emphasised the advantages of integrating diverse data types from electronic health records (EHRs) to enhance prediction model accuracy and minimise errors. Their model specifically combined structured data with unstructured text, suggesting that including additional data modalities like image data could be beneficial, especially if each modality contributes unique yet complementary information. The continuous evolution and innovation in machine learning techniques for healthcare applications indicate a promising future in enhancing patient care and hospital management through accurate predictive models.

This study devised a hybrid deep-learning model to predict patients' length of stay (LoS) within a hospital setting. The hybrid model amalgamates the strengths of three distinct learning frameworks: multi-layer convolutional neural networks, gated recurrent units, and dense neural models. Additionally, a fast hierarchical grid search technique was formulated to fine-tune the hyper-parameters of the hybrid model efficiently. This hyper-parameter optimisation tool proves swift and proficient in identifying a nearly optimal or locally optimal configuration within a minimal number of iterations. Leveraging a vast and comprehensive dataset comprising over 2.3 million records sourced from the New York State Department of Health [18], the Statewide Planning and Research Cooperative System (SPARCS) Inpatient De-identified file provided crucial discharge-level insights regarding patient attributes, diagnoses, treatments, services, and charges. Comparative analyses against ten machine learning and hybrid deep learning models showcased the superior performance of the proposed hybrid model, indicating meaningful advancements in LoS prediction accuracy.

## 2. Datasets and statistical analysis

In our research inquiry, we harnessed the openly accessible health data made available by the New York State SPARCS infrastructure [18]. The dataset under scrutiny pertained to 2021, representing the latest information accessible during our investigation. Comprising a voluminous CSV file, the dataset encapsulated a substantial array of 2.34 million rows and 34 columns. Each row within this dataset encapsulated anonymised data related to inpatient discharges, offering a rich tapestry of insights. Within the dataset columns, a diverse spectrum of information unfolded. These encompass geographic indicators linked to the caregiving institution, demographic markers like patient ethnicity, race, and age, as well as medical attributes such as the CCS diagnosis code, APR DRG code, illness severity metrics, and duration of hospital stay. Furthermore, payment facets were intricately woven into the dataset, detailing specifics like insurance types, overall charges, and procedural costs.

A detailed summary can be referenced to elucidate the dataset specifics [7] comprehensively. Descriptions of the CCS diagnosis code have been previously expounded. Embracing a wide spectrum of patients who underwent inpatient procedures across all New York State hospitals [19], the dataset delineated a panorama of payment sources. The dataset, stemming from the New York State SPARCS framework, not only embraces a diverse patient cohort extending beyond Medicare/Medicaid beneficiaries but also offers heightened value when mixed against datasets confined solely to Medicare/Medicaid demographics. Table 1 lists the feature details of the dataset and their types, numerical or categorical, and the group's number.

Table 1: The key features of admissions and patients information dataset

| Features | Type | Group | Features | Type | Group |
|---|---|---|---|---|---|
| Hospital Service Area | Categorical | 8 groups | Length Of Stay | Numerical | 0 to 140 |
| Hospital County | Categorical | 57 groups | Patient Disposition | Categorical | 19 groups |
| AgeGroup | Categorical | 5 groups | CCSR Diagnosis Code | Categorical | 471 groups |
| ZipCode 3Digits | Numerical | 100 to 149 | CCSR Diagnosis Description | Categorical | 472 groups |
| Gender | Categorical | 3 groups | APR DRG Code | Categorical | 326 groups |
| Race | Categorical | 4 groups | APR DRG Description | Categorical | 326 groups |
| Ethnicity | Categorical | 4 groups | APR MDC Code | Categorical | 24 groups |
| Type Of Admission | Categorical | 6 groups | APR MDC Description | Categorical | 24 groups |
| APR Severity Of Illness Code | Categorical | 4 groups | APR Risk Of Mortality | Categorical | 4 groups |
| APR Severity Of Illness Description | Categorical | 4 groups | APR Medical Surgical Description | Categorical | 2 groups |
| Payment Typology 1 | Categorical | 9 groups | Emergency Department Indicator | Logical | 2 options |
| Total Costs | Neumerical | $100 to 2 × 10^5 | Hospital County | Categorical | 57 groups |

## 2.1. Data wrangling

During the data wrangling process for the ED visitors dataset, our primary objective was to prepare the data for predicting the Length of Stay (LoS) in the hospital. This dataset encompasses crucial variables such as Hospital Service Area, Hospital County, age group, ZipCode 3Digits, gender, race, Ethnicity, Type Of Admission, APR Severity Of Illness Code, APR Severity Of Illness Description, Payment Typology, Total Costs, and the duration of hospitalisation. We accurately addressed missing values, rectified data formats and standardised variables, and ensured data uniformity. With a low rate of missing values, we employed a K-Nearest Neighbors (KNN) imputation method to fill in the gaps. We leveraged neighbouring values in key variables like age group, gender, race, and ethnicity to uphold dataset integrity. Categorical variables were encoded into numerical formats for statistical analysis, and as part of feature engineering, we extracted additional features from the admission date, including weekday, month, and year. To promote fair feature contribution in analysis and model training, we normalised the data by scaling numerical variables to a standardised range between 0 and 1. This comprehensive data-wrangling approach sets a solid foundation for accurate LoS prediction models and insightful analyses in the healthcare domain.

## 2.2. Statistical analysis

We analysed the feature distribution within the dataset, as illustrated in Figure 1. Notably, females accounted for a higher percentage than males, approximately 9%. Regarding the APR medical-surgical descriptions, the admission rate for medical cases was three times that of surgical cases. The predominant visitor demographic comprised seniors aged 70 years and older, exceeding the count of patients in the 18 to 29 age group by threefold and surpassing those under 17 years by twofold. It is evident that emergency-related admissions constituted the highest proportion, with over 1.3 million recorded instances.

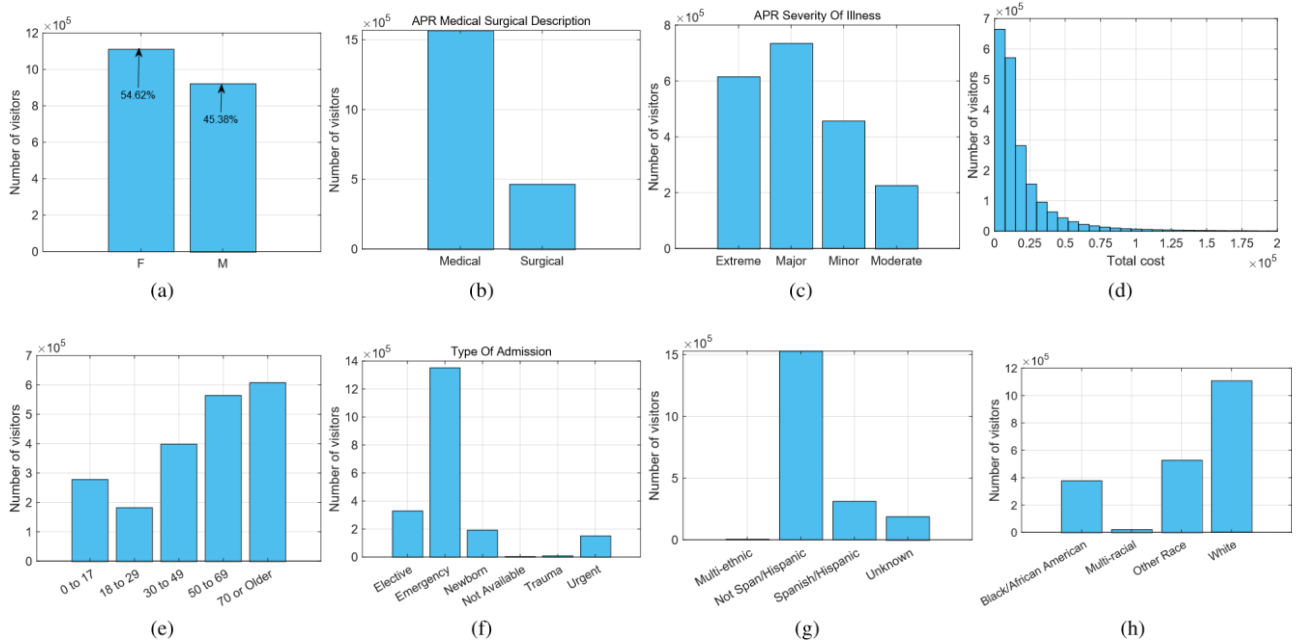

Figure 1: The distribution of the admissions and patients features information dataset. a) gender, b) APR medical surgical descriptions, c) APR severity of illness, d) Total cost of staying at the hospital, e) Age groups, f) Type of admission, g) Ethnicity, h) Race.

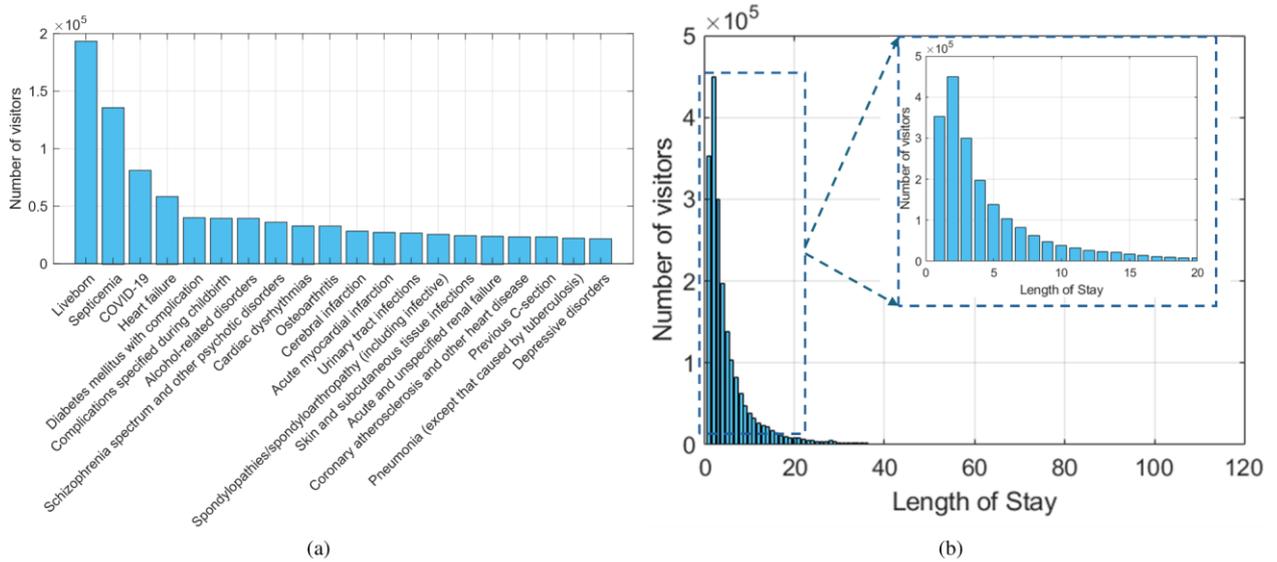

Figure 2: (a) The 20 most frequent diagnoses (ICD-10) at the hospital, (b) the distribution of stay length of patients at the hospital.

The distribution of the 20 most frequent diagnoses utilising the ICD-10 system is depicted and organised in Figure 2 (a). The primary cause for admission, with the highest frequency at $1.9 \times 10^5$, was 'Single liveborn infant.' Following, 'Septicemia,' 'COVID-19,' and 'Heart failure' occupied the second to fourth positions among the top 20 ICD-10 codes. Figure 2 (b) illustrates the distribution concerning the length of stay, emphasising a notable concentration within the category of less than 20 days. Only a minor fraction of patients, constituting 4%, extended their stay beyond the 20-day mark.

To assess the linear correlation between features and Length of Stay (LoS), we calculated the R-value, with the outcomes visualised in Figure 3. As anticipated, the most significant correlation between hospitalisation cost and LoS was observed, exceeding 60%. Additionally, the APR severity of illness code exhibited the second-highest correlation with LoS, underscoring its pivotal role in predicting the Length of Stay.

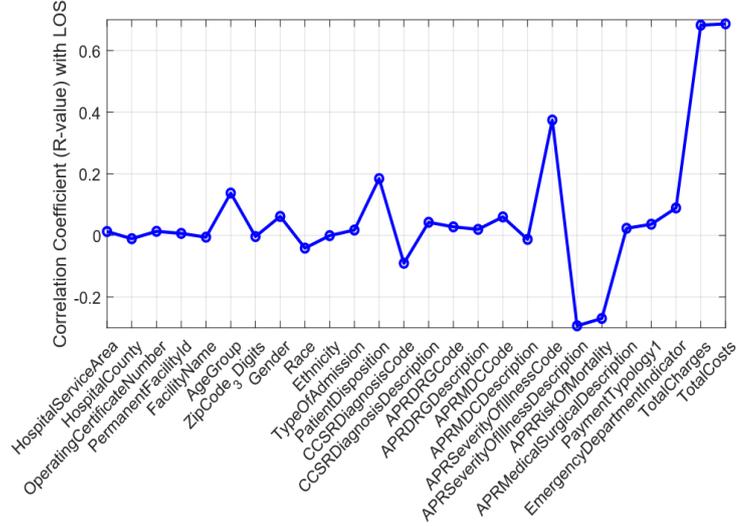

Figure 3: The correlation coefficient of Length of patients' stay in hospital with other admission information.

## 3. Methods

One of the robust and reliable types of ML models is Recurrent neural networks (RNNs), which operate on sequential data in a unique manner [20], handling one element at a time while also selectively transferring information across various steps within the sequence. This distinctive characteristic enables RNNs to effectively model input and/or output sequences containing interdependent elements rather than isolated entities [21]. RNNs showcase their prowess by adeptly capturing not only sequential patterns but also temporal dependencies across different scales, all at the same time. Their remarkable flexibility in grasping contextual cues and temporal correlations renders them exceptionally valuable in a wide array of applications, such as natural language processing, time series forecasting, and speech recognition.

### 3.1. GRU: Gated Recurrent Units

One of the effective and popular Recurrent Neural Networks (RNNs) is the Gated Recurrent Unit (GRU) network [22]. It tackles the challenge of capturing long-range dependencies within sequential data through its adept use of gating mechanisms. These mechanisms enable the network to selectively update its hidden state at each time step, enhancing its ability to retain essential information while discarding irrelevant details. Specifically, one crucial component of GRU networks is the Update Gate. This pivotal element plays a decisive role in determining the flow of information within the network, guiding what information is propagated forward and what is deemed unnecessary for the current context.

$$Z_t = \sigma_g \, W_z x_t + U_z h_{t-1} + b_z \qquad (1)$$

In the context provided, the update gate denoted as $Z_t$, is influenced by weight matrices $W_z, U_z$, and bias term $b_z$, with the sigmoid activation function $\sigma_g$ playing a pivotal role in determining its behaviour. This mechanism allows for the selective updating of information within the network, crucial for managing the flow of data at each time step. Moving on to the reset gate, also known as the barrier gate, this component is tasked with resetting the hidden state to zero at specific junctures in the sequence, enriching the network's ability to adapt to changing contexts effectively as follows.

$$r_t = \sigma_g\, W_{rx_t} + U_r h_{t-1} + b_r \qquad (2)$$

The candidate hidden state, $h^{(.)}_t$, is the intermediate memory state before the update. It is defined as follows:

$$h_{(t.)} = \tanh W_{hx_t} + U_h h_{t-1} + b_h \qquad (3)$$

Finally, the hidden state, $h_t$, is updated using the update gate and the candidate hidden state as follows:

$$h_t = (1 - z_t) \odot h_{t-1} + z_t \odot h^{(.)}_t \qquad (4)$$

### 3.2. CNNs: Convolutional Neural Networks

CNN has demonstrated its reliability in extracting hidden features by autonomously generating filters [23]. Within a CNN's convolutional layer, neurons operate autonomously without direct connections, sharing filter weights. This unique architecture enables CNN to outperform MLP in training efficiency when employing equivalent layers and neurons. Each convolutional layer can be visualized as a self-contained feature extractor, underscoring CNN's effectiveness in processing complex data patterns.

$$h^k_{ij} = f\big((W^k * x)_{ij} + b_k\big) \qquad (5)$$

where $f$ is related to the activation function, $W^k$ denotes the weights of the kernel connected to the $k^{\text{th}}$ feature map. To enhance representational capacity, nonlinear activation functions such as the hyperbolic tangent, sigmoid, and rectified linear unit (ReLU) are typically utilised at the output of each layer. ReLU has emerged as a prevalent choice for improving network trainability. The formula for ReLU activation is as follows: $a^{l(i,j)} = f\big(y^{l(i,j)}\big) = \max\big(0, y^{l(i,j)}\big)$, where $a^{l(i,j)}$ is the activation output of $y^{(li,j)}$.

### 3.3. Proposed hybrid deep learning model

The primary limitation of fully connected RNN models when handling spatiotemporal data lies in their heavy reliance on complete connections for input-to-state and state-to-state transitions, as highlighted in [21]. This approach neglects the incorporation of spatial context, treating all input features uniformly without considering the spatial relationships among different components in a sequence. This poses challenges in tabular datasets such as those related to LoS, where spatial connections are less explicit, making it challenging for CNNs to extract relevant features. Furthermore, LoS datasets often comprise diverse features with intricate interactions and dependencies, which CNNs may not effectively capture in comparison to other architectures.

To tackle the outlined challenges, we have introduced a novel hybrid deep learning methodology that integrates multi-layer CNNs, GRU, and dense neural networks to leverage each component's strengths while mitigating their limitations. Initially, we conducted a comprehensive evaluation of three key sequential learning models - LSTM, BiLSTM, and GRU - due to their distinct capabilities in handling nonlinear relationships within features and target modelling. Subsequently, we constructed a stacked model based on the most effective model identified in the initial phase, fine-tuning the layer count through a greedy search strategy. This stacked model was then combined with a multi-layer CNN to harness the complementary advantages of both architectures. Lastly, incorporating a series of fully connected layers into the hybrid model further enhanced its predictive performance. Figure 4 illustrates the sequential steps and constituent elements of our proposed hybrid model tailored for predicting patients' length of stay in hospital settings.

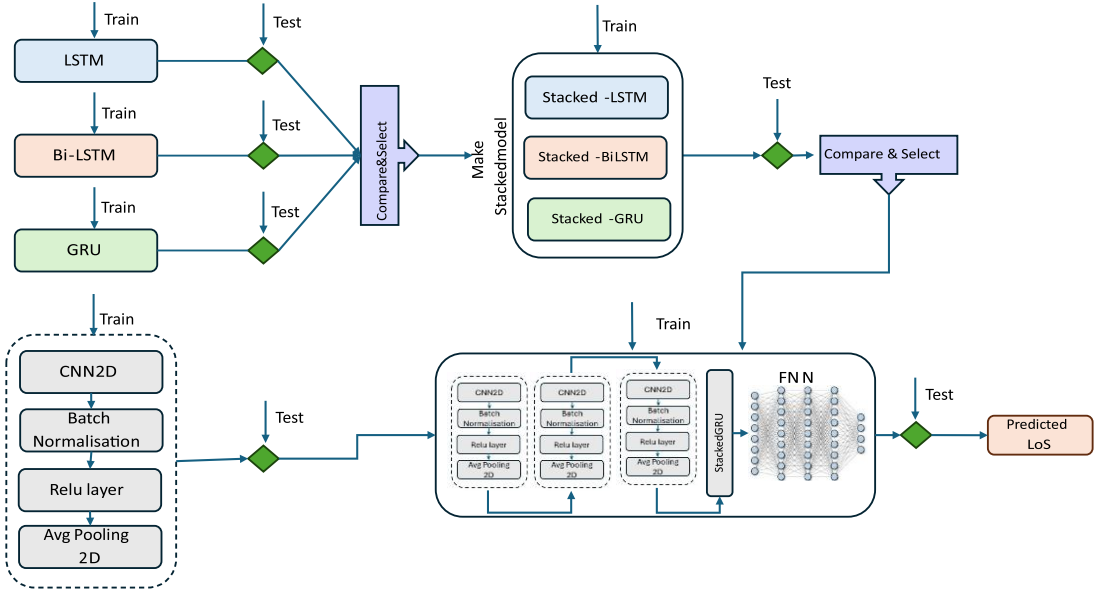

Figure 4: Schematic model of the proposed hybrid deep learning model for predicting the LoS.

## 4. Experimental modelling results

In this section, we delve into the findings of length of stay (LoS) prediction utilising the proposed hybrid deep learning model, compared with 11 alternative prediction models, employing five distinct performance metrics for evaluation. These metrics include Mean Squared Error (MSE), Root Mean Squared Error (RMSE), Loss function, Mean Absolute Error (MAE), and the coefficient of determination (R-value). A higher R-value signifies superior modelling accuracy, indicating proximity between predicted and actual LoS values. Conversely, lower values indicate enhanced predictive performance for error-based metrics, such as MSE, RMSE, and MAE. Table 2 delineates the formulations of the aforementioned metrics applied within this study to assess the efficacy of the predictive models.

Table 2: ML and hybrid deep learning models' performance metrics and formulations.

| Metric | Equation |
|---|---|
| Mean Squared Error (MSE) | $= \frac{1}{N} \sum_{i=1}^{N} (y_i - \hat{y})^2,$ |
| Root Mean Squared Error (RMSE) | $= \sqrt{\frac{1}{N} \sum_{i=1}^{N} (y_i - \hat{y})^2}$ |
| Loss | $= \frac{1}{2N} \sum_{i=1}^{N} (y_i - \hat{y})^2$ |
| Mean absolute error (MAE) | $= \frac{1}{N} \sum_{i=1}^{N} |y_i - \hat{y}|$ |
| Coefficient of Determination ($R$) | $= 1 - \frac{\sum (y_i - \hat{y})^2}{\sum (y_i - \bar{y})^2},$ |

### 4.1. Modelling results

In our investigation of the proposed hybrid learning model's efficacy in forecasting the length of stay (LoS) for hospitalized patients, a comprehensive comparative framework was devised, comprising three sequential neural networks—LSTM, Bi-LSTM, GRU—alongside a multi-layer CNN. Additionally, the framework incorporated three stacked models derived from the sequential networks and three hybrid models created by integrating the stacked models with CNN. The statistical outcomes of this assessment are delineated in Table 3, showcasing the peak LoS accuracy achieved by the proposed model (CNN-GRU-DNN) at an average of 0.89 across a 10-fold cross-validation test. Subsequently, a fusion of stacked GRU and multi-layer CNN attained an accuracy of 0.87. The results reaffirm the advantage of CNN on LoS prediction accuracy, facilitating the model in harnessing spatial and temporal information within the tabular dataset. This holistic approach enhances the precision of LoS predictions by considering a broader spectrum of factors influencing patient outcomes. Notably, in terms of mean absolute error (MAE) during testing, the CNN-BiLSTM model exhibited the lowest error at 2.37 days of LoS. Nevertheless, across other testing error metrics like MSE, RMSE, and Loss, the proposed model (CNN-GRU-DNN) outperformed alternative prediction models with an RMSE of 3.99 days. Moreover, the p-values obtained (See Table 3) from the statistical T-test comparing the performance (R-value) of 11 different models with our proposed model are remarkably low (< 0.05). Such tiny p-values strongly indicate significant differences in the predictive performance between these models and your proposed one.

Table 3: Statistical results of 11 LoS prediction models compared with the proposed hybrid model based on five evaluation metrics. $p$ shows the p-value

| | LSTM ($p = 6.8e - 36$) | | | | | Bi-LSTM ($p = 1.6e - 39$) | | | | |
|---|---|---|---|---|---|---|---|---|---|---|
| Metric | MSE | RMSE | LOSS | MAE | R | MSE | RMSE | LOSS | MAE | R |
| Mean | 28.231 | 5.313 | 14.116 | 2.902 | 0.748 | 27.796 | 5.272 | 13.898 | 2.856 | 0.751 |
| Max | 29.206 | 5.404 | 14.603 | 2.972 | 0.755 | 28.483 | 5.337 | 14.242 | 2.915 | 0.759 |
| Min | 27.343 | 5.229 | 13.672 | 2.824 | 0.734 | 27.447 | 5.239 | 13.724 | 2.792 | 0.741 |
| STD | 0.688 | 0.065 | 0.344 | 0.050 | 0.008 | 0.369 | 0.035 | 0.185 | 0.037 | 0.005 |
| | GRU ($p = 2.7e - 37$) | | | | | CNN ($p = 2.5e - 35$) | | | | |
| Metric | MSE | RMSE | LOSS | MAE | R | MSE | RMSE | LOSS | MAE | R |
| Mean | 27.865 | 5.279 | 13.932 | 2.828 | 0.749 | 24.216 | 4.912 | 12.108 | 2.976 | 0.831 |
| Max | 28.407 | 5.330 | 14.204 | 2.892 | 0.758 | 29.219 | 5.405 | 14.609 | 3.536 | 0.847 |
| Min | 26.979 | 5.194 | 13.489 | 2.773 | 0.739 | 21.006 | 4.583 | 10.503 | 2.638 | 0.816 |
| STD | 0.430 | 0.041 | 0.215 | 0.036 | 0.007 | 3.146 | 0.316 | 1.573 | 0.283 | 0.010 |
| | S-LSTM ($p = 7.5e - 35$) | | | | | S-BiLSTM ($p = 1.6e - 34$) | | | | |
| Metric | MSE | RMSE | LOSS | MAE | R | MSE | RMSE | LOSS | MAE | R |
| Mean | 25.277 | 5.027 | 12.639 | 2.643 | 0.766 | 27.796 | 5.272 | 13.898 | 2.856 | 0.751 |
| Max | 25.935 | 5.093 | 12.967 | 2.701 | 0.776 | 28.483 | 5.337 | 14.242 | 2.915 | 0.759 |
| Min | 24.539 | 4.954 | 12.269 | 2.593 | 0.756 | 27.447 | 5.239 | 13.724 | 2.792 | 0.741 |
| STD | 0.416 | 0.041 | 0.208 | 0.031 | 0.007 | 0.369 | 0.035 | 0.185 | 0.037 | 0.005 |
| | S-GRU ($p = 3.6e - 18$) | | | | | CNN-LSTM ($p = 4.1e - 4$) | | | | |
| Metric | MSE | RMSE | LOSS | MAE | R | MSE | RMSE | LOSS | MAE | R |
| Mean | 27.865 | 5.279 | 13.932 | 2.828 | 0.749 | 18.097 | 4.251 | 9.048 | 2.588 | 0.869 |
| Max | 28.407 | 5.330 | 14.204 | 2.892 | 0.758 | 20.151 | 4.489 | 10.076 | 2.972 | 0.874 |
| Min | 26.979 | 5.194 | 13.489 | 2.773 | 0.739 | 16.121 | 4.015 | 8.060 | 2.329 | 0.857 |
| STD | 0.430 | 0.041 | 0.215 | 0.036 | 0.007 | 1.444 | 0.170 | 0.722 | 0.229 | 0.005 |
| | CNN-BiLSTM ($p = 1.6e - 09$) | | | | | GRU-CNN ($p = 8.7e - 12$) | | | | |
| Metric | MSE | RMSE | LOSS | MAE | R | MSE | RMSE | LOSS | MAE | R |
| Mean | 16.701 | 4.086 | 8.351 | 2.365 | 0.871 | 17.045 | 4.127 | 8.522 | 2.515 | 0.874 |
| Max | 17.447 | 4.177 | 8.723 | 2.473 | 0.875 | 18.770 | 4.332 | 9.385 | 2.839 | 0.878 |
| Min | 15.481 | 3.935 | 7.740 | 2.163 | 0.865 | 15.950 | 3.994 | 7.975 | 2.316 | 0.868 |
| STD | 0.760 | 0.094 | 0.380 | 0.109 | 0.003 | 0.879 | 0.106 | 0.439 | 0.159 | 0.003 |
| | CNN-GRU-DNN ($p = 7.5e - 11$) | | | | | CNN-GRU-DNN-S | | | | |
| Metric | MSE | RMSE | LOSS | MAE | R | MSE | RMSE | LOSS | MAE | R |
| Mean | 15.996 | 3.993 | 7.998 | 2.415 | 0.888 | 17.834 | 4.217 | 8.917 | 2.517 | 0.867 |
| Max | 19.825 | 4.453 | 9.913 | 2.858 | 0.893 | 19.706 | 4.439 | 9.853 | 2.774 | 0.869 |
| Min | 13.576 | 3.685 | 6.788 | 2.151 | 0.884 | 15.961 | 3.995 | 7.981 | 2.260 | 0.864 |
| STD | 1.932 | 0.238 | 0.966 | 0.233 | 0.003 | 2.648 | 0.314 | 1.324 | 0.363 | 0.004 |

Figure 5 presents a detailed analysis of the statistical results, including the minimum, maximum, median, and outliers, from a 10-fold cross-validation evaluation of the proposed hybrid deep learning model alongside 11 other models, focusing on accuracy (R-value) and RMSE metrics. As can be seen, the proposed model outperformed its counterparts, achieving a median R-value exceeding 0.88, marking a significant milestone. Despite attempts to enhance the CNN-GRU-DNN architecture by integrating a self-attention mechanism, the results revealed a lack of performance improvement. This outcome was attributed to the potential for overfitting, which could compromise the model's generalisation capabilities, ultimately leading to a decline in performance. Parallel observations from the RMSE comparisons in Figure 5(b) reaffirmed the superiority of the proposed model over others. Interestingly, while models like LSTM, GRU, BiLSTM, and their stacked versions exhibited suboptimal performance, particularly in terms of RMSE, their performance variance remained narrower compared to the hybrid model, suggesting potential robustness in specific scenarios.

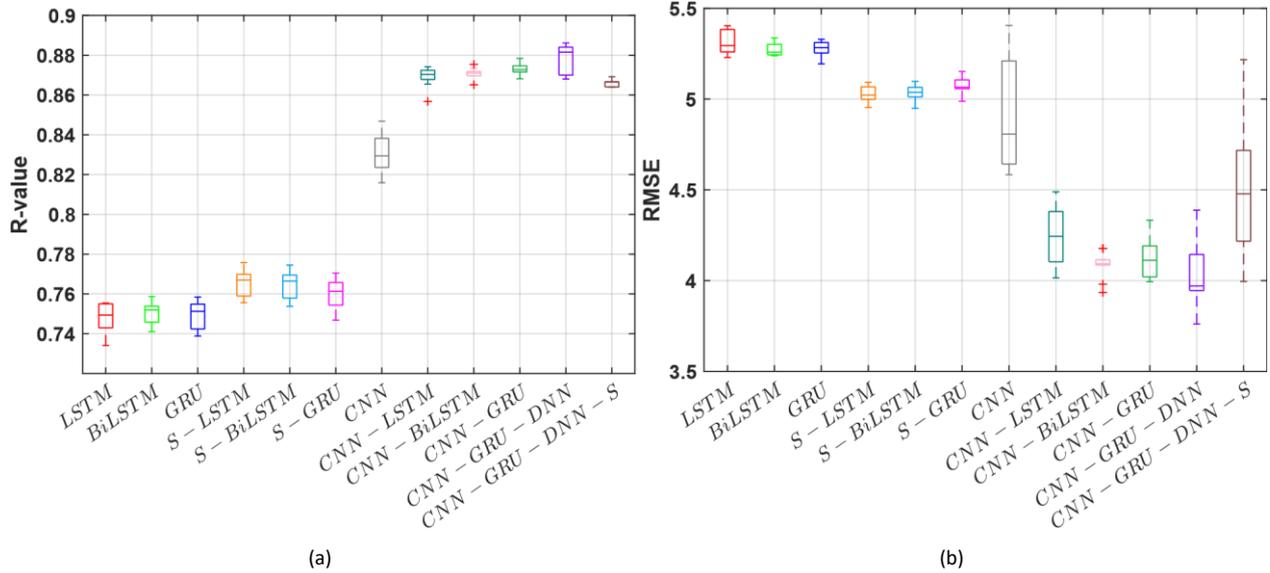

(a)                                                            (b)

Figure 5: Statistical analysis of the proposed hybrid deep learning model (CNN-GRU-DNN) compared with other 11 models based on the average of 10-fold cross validation (a) accuracy (R-value), and (b) RMSE of the test.

## 4.2. Feature selection

In this study, we developed a simple feature selection procedure to investigate the significance of each attribute in enhancing the efficacy of the proposed hybrid model. Initially trained on the complete feature set, the model underwent a systematic elimination process to measure the impact of each feature on performance metrics such as accuracy, Mean Absolute Error (MAE), and Root Mean Squared Error (RMSE). This iterative process continued until each feature was eliminated at least once. Notably, the results illustrated in Figure 6 highlight the superior performance gains achieved by discarding specific attributes like hospital country, race, and Emergency Department (ED) indicator in terms of accuracy enhancement. Besides, as depicted in Figure 6(b), features such as Hospital service area, Hospital country, Zipcode, Race, and APR MDC description emerge as potential candidates for removal to optimize MAE performance within the model.

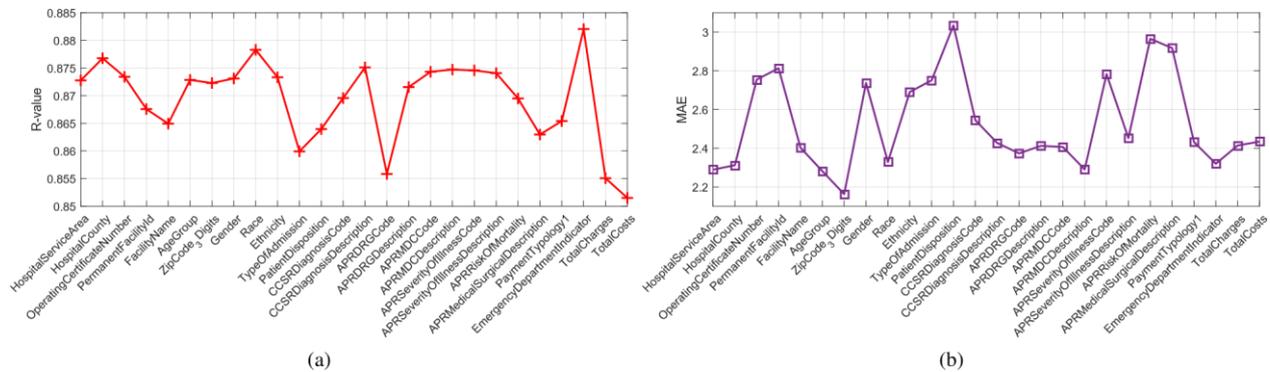

(a)                                                            (b)

Figure 6: Feature selection based on the average of 10-fold cross validation (a) accuracy (R-value), and (b) MAE of the test.

*4.3. Hyper-parameters tuning*

Exploring various settings is crucial to constructing an efficient deep-learning model. Hyper-parameter tuning [24], vital in crafting effective models, is particularly significant for large and deep models, which boast numerous hyper-parameters. Unlike manual tuning, which demands in-depth algorithmic comprehension, automatic techniques like hyper-parameter optimisation (HPO) are gaining traction for their ability to streamline and enhance the tuning process. HPO aims to automate hyper-parameter tuning, simplifying the application of the models to real-world challenges, ultimately culminating in an optimised model architecture.

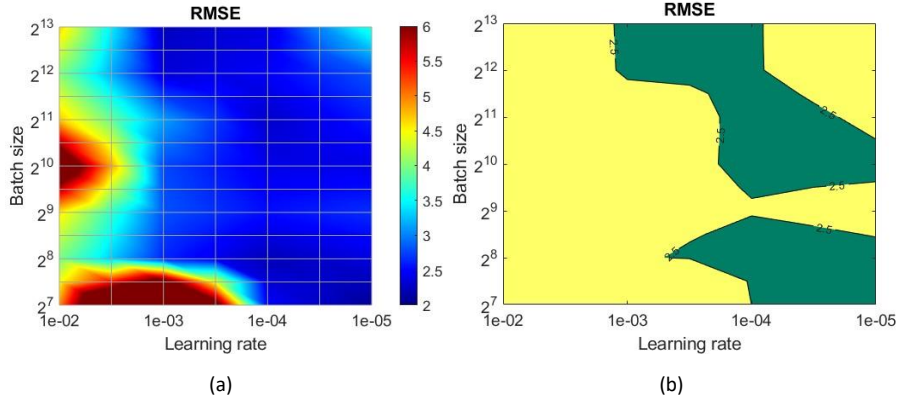

Figure 7: Hyper-parameter tuning for CNN-GRU-DNN using grid search focused on batch size and learning rate.

In our exploration, we concentrated on the critical hyper-parameters of batch size and learning rate. They are critical in deep learning models that influence training efficiency, convergence speed, generalisation ability, and overall model performance. Finding the right balance for these parameters is essential for achieving optimal results in training deep neural networks. We accurately examined various combinations of these parameters using a grid search approach. We selected four learning rate values ranging from $1e-02$ to $1e-05$ and seven batch size values from $2^7$ to $2^{13}$. This method enabled us to train and test our hybrid model consistently with identical data and random seeds, ensuring a fair assessment of the predicted LoS outcomes. Figure 7 illustrates the grid search landscape depicting the model's performance based on the test data's RMSE. Notably, the RMSE landscape unveils two distinct clusters representing the optimal learning rate and batch size configurations. A key insight gleaned from this hyper-parameter tuning analysis is the prominence of the learning rate over batch size. This observation underscores that a smaller learning rate can yield superior predictive outcomes while maintaining the same iteration count.

## 5. Conclusion

The fusion of cutting-edge technology with the intricacies of healthcare management has unveiled a transformative path towards precision-driven patient care. Predicting hospital length of stay (LoS) has emerged as a pivotal cog in the machinery of public health strategies, illuminating trends and patterns that redefine the contours of healthcare delivery. Our pioneering venture into predictive analytics introduces an effective hybrid deep learning model, a mixture of Multi-layer Convolutional Neural Networks (CNNs), Gated Recurrent Units (GRU), and dense neural networks. This model produced a high accuracy and outperformed conventional and state-of-the-art Machine Learning (ML) and Deep Learning (DL) models.

Through a detailed analysis of diverse variables encompassing geographic nuances, demographic intricacies, and medical specifics, our model emerges as a beacon of precision in the labyrinth of healthcare analytics. The statistical analysis achieved by our CNN-GRU-DNN model, soaring at an average LoS accuracy of 89% across a rigorous 10fold cross-validation, exceeds benchmarks like LSTM, BiLSTM, GRU, and CNNs by remarkable margins of 19%, 18.2 18.6%, and 7%, respectively. This level of precision infuses confidence in the model's predictive capabilities, reassuring healthcare professionals and administrators of its potential impact.

The accurate LoS prophecies not only empower hospitals to recalibrate resource distribution and mitigate the fiscal strains of prolonged stays but also unveil uncharted avenues in hospital stay governance. This trajectory of innovation promises to catalyse healthcare research and innovation and also to sculpt a future where precision-driven healthcare practices sculpt a narrative of holistic well-being and patient-centric care.